\documentclass[11pt]{article}

\usepackage[margin=1in]{geometry}
\usepackage[T1]{fontenc}
\usepackage{lmodern}
\usepackage{microtype}
\usepackage{amsmath,amssymb,amsfonts}
\usepackage{graphicx}
\usepackage{booktabs}
\usepackage{float}
\usepackage{placeins}
\usepackage[hidelinks]{hyperref}
\usepackage{authblk}

\title{When Regulation Has Memory:\\
Hysteresis and Control Burden in Artificial Agency}

\author[1]{Veronique Ziegler}
\affil[1]{Independent Researcher}
\date{}

\begin{document}
\maketitle

\begin{abstract}
Adaptive agents are usually judged by what they do, but an agent can appear stable while the internal effort required to keep it stable is increasing. This hidden regulatory burden matters for artificial agents operating under noise, delay, or changing demands: two systems may reach similar internal states while one requires much more corrective control to get there.

Here, we study whether that burden depends on history. Using a computational model of adaptive uncertainty regulation, we drive an artificial agent through a continuous change in its uncertainty target and then reverse the change without resetting the agent. This creates a simple test for carryover: does the controller respond only to the current target, or does the path by which the agent reached that target still matter?

The simulations show a clear history-dependent effect. The adaptive gain required to regulate the agent forms a reproducible hysteresis loop, meaning that the same target can require different levels of control depending on whether the agent is moving toward or returning from a more demanding regime. The timing of regulation also matters. When stabilization is available before disturbance exposure, the agent generally requires less adaptive gain than when it can only recover after disturbance has already acted.
 
The state-level coherence measure also shows path dependence, but the timing effect is much clearer in regulatory gain.
The main difference is therefore not that anticipatory regulation produces a completely different state. Rather, it reaches comparable regulated behavior with lower modeled control demand. These results suggest that adaptive agents should be evaluated not only by whether they remain organized, but by how much regulation they must recruit to do so.

\end{abstract}

\section{Introduction}

Adaptive regulation is often evaluated by whether an agent remains stable under disturbance. A separate question is how much regulatory effort is required to maintain that stability, and whether this effort depends on the agent's prior trajectory. This distinction matters because two agents may reach similar regulated states while carrying different histories of control demand.
To study this, we use IRAM-$\Omega$-Q, a framework previously introduced for modeling uncertainty regulation in adaptive artificial agents \cite{ziegler2026iramomegaq}.  That work showed that causal ordering can affect regulatory demand under fixed operating conditions. 
Here, we test whether regulatory demand is path-dependent: whether the amount of stabilization required at a given target depends on the trajectory by which the agent reached that target.

 In many adaptive systems, the control required at a given target may depend not only on the current target value, but also on the path by which the system arrived there. A system being driven toward a more demanding condition may recruit regulation differently from a system returning from that condition, even when the instantaneous target is the same.
This kind of carryover can be tested with a hysteresis protocol. The target is varied continuously upward and then downward while the internal state and controller are not reset. If the required regulation depends only on the instantaneous target, the upward and downward branches should overlap. If the system retains history, the return branch can differ from the outward branch.
This use of hysteresis is operational: branch separation is treated as evidence of retained trajectory history, consistent with broader nonlinear-dynamical accounts of path dependence and self-organizing behavior \cite{kelso1995dynamic,strogatz2015nonlinear}.

We apply this protocol to an artificial agent simulated within IRAM-$\Omega$-Q.
The target entropy is ramped from $S^\ast=0.15$ to $0.45$ and then back to $0.15$ in a single continuous trajectory. Regulation-first and disturbance-first orderings are compared using matched stochastic seeds. The central question is whether the causal-ordering effect observed under fixed conditions persists when the simulated agent is driven through path-dependent dynamics.

The primary outcome is the adaptive gain $\mu(t)$, interpreted as model-level regulatory demand. The state-level outcome is the coherence gap $\Delta C$, interpreted as a diagnostic of retained internal organization. This distinction is important: two trajectories may show broadly comparable state-level behavior while requiring different amounts of adaptive regulation to sustain it.

In this paper, we show that the simulated agent exhibits hysteresis under continuous target modulation: the adaptive gain required at a given target depends on the prior trajectory. We also show that regulation-first timing reduces the regulatory burden required to traverse this retained-history protocol.

\section{Model definitions}

The full IRAM-$\Omega$-Q architecture is described in earlier work \cite{ziegler2026iramomegaq}. The present study uses the same state representation, entropy diagnostics, adaptive-gain controller, and regulation/disturbance orderings. This section summarizes only the definitions needed to interpret the continuous target-ramp protocol.

The evolving internal state is a normalized complex amplitude vector,
\begin{equation}
\psi(t) = \bigl(\psi_1(t), \ldots, \psi_d(t)\bigr),
\qquad
\sum_i |\psi_i(t)|^2 = 1 .
\end{equation}
At each measurement step, a density matrix $\rho(t)$ is constructed from the current state and used to compute the controller entropy and coherence-gap diagnostic.

The controller uses the von Neumann entropy,
\begin{equation}
S_{\mathrm{vN}}(\rho)
=
-\mathrm{Tr}(\rho\log\rho)
=
-\sum_k \lambda_k \log \lambda_k ,
\end{equation}
as its internal uncertainty signal. The diagonal entropy is
\begin{equation}
S_{\mathrm{diag}}(\rho)
=
-\sum_i \rho_{ii}\log\rho_{ii},
\end{equation}
and the coherence gap is
\begin{equation}
\Delta C = S_{\mathrm{diag}} - S_{\mathrm{vN}} .
\end{equation}

The adaptive gain $\mu(t)$ controls the strength of stabilization and is interpreted here as a model-level measure of regulatory demand. The implemented gain update is
\begin{equation}
\mu(t+\Delta t)
=
\Pi_{[\mu_{\min},\mu_{\max}]}
\left[
\mu(t)
+
\alpha\,\dot S_{\mathrm{vN}}(t)
+
\beta\bigl(S_{\mathrm{vN}}(t)-S^\ast(t)\bigr)
\right],
\label{eq:mu-update}
\end{equation}
where $\Pi_{[\mu_{\min},\mu_{\max}]}$ clips the result to the admissible gain range. Under this convention, entropy above the target contributes positively to the gain, so that the controller recruits stronger stabilization when uncertainty exceeds the target level.
This use of an adaptive gain follows the control-theoretic idea that feedback strength can be adjusted in response to deviations from a target state \cite{astrom2008feedback}.

The coherent part of the dynamics is generated by a Hamiltonian $H$ and applied with the exact finite-time propagator
\begin{equation}
\psi(t+\Delta t)=e^{-iH\Delta t}\psi_{\mathrm{nc}}(t),
\end{equation}
where $\psi_{\mathrm{nc}}(t)$ is the state after the non-coherent disturbance and regulation operations for that time step. This coherent update is the same in both causal orderings.

The two causal orderings differ only in the placement of regulation relative to incoming disturbance. In regulation-first control, the updated gain is available before current-cycle disturbance exposure:
\begin{equation}
\psi(t)
\xrightarrow{\;S_{\mathrm{vN}},\,\mu^+\;}
\mathcal{M}_{\mu^+}\psi(t)
\xrightarrow{\;\eta_{\mathrm{eff}}=\eta(1-\mu^+)\;}
\psi_{\mathrm{nc}}^{\mathrm{RF}}(t).
\end{equation}
Here $\mathcal{M}_{\mu}$ denotes the stabilization operation, $\mu^+$ denotes the updated gain, and $\eta$ is the incoming disturbance amplitude.

In disturbance-first control, incoming disturbance acts before the new regulatory response is computed:
\begin{equation}
\psi(t)
\xrightarrow{\;\eta\;}
\psi_{\eta}(t)
\xrightarrow{\;S_{\mathrm{vN}},\,\mu^+\;}
\mathcal{M}_{\mu^+}\psi_{\eta}(t)
=
\psi_{\mathrm{nc}}^{\mathrm{DF}}(t).
\end{equation}
The externally specified disturbance amplitude is the same in both cases. The difference is the timing of stabilization relative to exposure.

\section{Methods}

\subsection{Continuous target-ramp protocol}

A hysteresis protocol requires a continuous trajectory: the downward branch must inherit the state and controller history produced on the upward branch. We therefore use a triangular target-entropy schedule in a single run,
\begin{equation}
S^\ast(t)=
\begin{cases}
S^\ast_{\min}+2x(t)\bigl(S^\ast_{\max}-S^\ast_{\min}\bigr), & x(t)\leq \tfrac{1}{2},\\[4pt]
S^\ast_{\max}+\bigl(2x(t)-1\bigr)\bigl(S^\ast_{\min}-S^\ast_{\max}\bigr), & x(t)>\tfrac{1}{2},
\end{cases}
\label{eq:target-ramp}
\end{equation}
where $x(t)$ is normalized simulation progress. State and controller values are not reset at the turning point.

The reference protocol uses $S^\ast_{\min}=0.15$, $S^\ast_{\max}=0.45$, initial gain $\mu_0=0.08$, incoming disturbance amplitude $\eta=0.13$, dimension $d=16$, and time step $\Delta t=0.01$. Each trajectory contains 30,000 steps. The main settings are summarized in Table~\ref{tab:protocol}.

\begin{table}[H]
\centering
\caption{\textbf{Continuous target-ramp protocol used for the matched RF/DF ensemble comparison.}}
\label{tab:protocol}
\begin{tabular}{ll}
\toprule
Parameter & Value \\
\midrule
Number of matched RF/DF replicate pairs & 30 \\
State dimension $d$ & 16 \\
Focus index & 6 \\
Time step $\Delta t$ & 0.01 \\
Steps per continuous trajectory & 30,000 \\
Incoming disturbance amplitude $\eta$ & 0.13 \\
Initial regulation gain $\mu_0$ & 0.08 \\
Target entropy schedule & triangular, continuous \\
Target entropy range & $0.15 \rightarrow 0.45 \rightarrow 0.15$ \\
Hamiltonian energy scale & 0.15 \\
Hamiltonian coupling $c$ & 0.08 \\
Interaction locality $\ell$ & 2.0 \\
Derivative feedback scale $\alpha$ & $5\times 10^{-4}$ \\
Target-error feedback scale $\beta$ & $2\times 10^{-4}$ \\
Gain bounds $(\mu_{\min},\mu_{\max})$ & $(0.001,1.0)$ \\
Base seed & 123456789 \\
\bottomrule
\end{tabular}
\end{table}

\subsection{Matched-seed ensemble design}

Thirty regulation-first trajectories and thirty disturbance-first trajectories were generated as matched pairs. Within each pair, the two trajectories used the same stochastic seed and identical numerical parameters. The causal ordering was the intended difference between conditions. This matched-seed design reduces variation caused solely by different random disturbance histories and makes the RF/DF comparison more directly interpretable as an ordering effect.

Each continuous trajectory was split into an increasing-target branch and a decreasing-target branch according to the target schedule. For each replicate, $\mu(t)$ and $\Delta C(t)$ were binned by $S^\ast$ separately on the two branches. Replicate-level binned curves were then averaged across the 30 matched pairs. Figure uncertainty bands show approximate 95\% confidence intervals computed as
\begin{equation}
\overline{y}(S^\ast) \pm 1.96\,\frac{s_y(S^\ast)}{\sqrt{N}},
\qquad N=30,
\end{equation}
where $y$ denotes either $\mu$ or $\Delta C$ and $s_y$ is the across-replicate standard deviation of branch-binned means.

\subsection{Paired hysteresis summaries}

In addition to ensemble branch curves, replicate-level signed hysteresis-loop areas are computed for each observable $y\in\{\mu,\Delta C\}$ and ordering $o\in\{\mathrm{RF},\mathrm{DF}\}$:
\begin{equation}
A_y^{(o)}
=
\int_{S^\ast_{\min}}^{S^\ast_{\max}}
\left[
y_{\downarrow}^{(o)}(S^\ast)-
y_{\uparrow}^{(o)}(S^\ast)
\right]\,dS^\ast .
\label{eq:area}
\end{equation}
Positive area means that the decreasing-target branch lies above the increasing-target branch, while negative area indicates the reverse loop orientation. Areas are evaluated independently for each matched-seed replicate using the same branch binning used for the ensemble trajectories.

To isolate the ordering effect directly, within-pair DF-minus-RF branch differences are also calculated:
\begin{equation}
D_y^{(b)}(S^\ast)
=
y_{\mathrm{DF}}^{(b)}(S^\ast)
-
y_{\mathrm{RF}}^{(b)}(S^\ast),
\qquad
b\in\{\uparrow,\downarrow\}.
\label{eq:paired-diff}
\end{equation}
The paired difference curves are averaged across matched seeds with approximate 95\% confidence intervals. For $y=\mu$, a positive value represents greater adaptive regulatory demand under disturbance-first control. For $y=\Delta C$, a nonzero value represents an ordering-dependent displacement in the state-level observable.

\subsection{Parameter-neighborhood robustness protocol}

The reference-condition analysis is extended to a local nine-condition neighborhood:
\begin{equation}
\eta \in \{0.08,0.13,0.18\},
\qquad
\mu_0 \in \{0.04,0.08,0.12\}.
\end{equation}
At every point of this $3\times3$ grid, the triangular target schedule remains fixed at $S^\ast\in[0.15,0.45]$. All Hamiltonian, focus-index, controller-feedback, time-step, and trajectory-length settings remain those of the reference protocol. Thirty matched-seed RF/DF trajectory pairs are generated per grid point. The complete robustness study therefore contains 270 matched pairs, or 540 simulated trajectories.

For each ordering and parameter cell, hysteresis is summarized by the replicate-averaged signed areas $A_\mu^{(o)}$ and $A_{\Delta C}^{(o)}$. Ordering-dependent burden is additionally summarized within each matched pair by
\begin{equation}
B_y
=
\left\langle
y_{\mathrm{DF}}^{(b)}(S^\ast)
-
y_{\mathrm{RF}}^{(b)}(S^\ast)
\right\rangle_{b\in\{\uparrow,\downarrow\},\,S^\ast},
\qquad
y\in\{\mu,\Delta C\}.
\label{eq:burden}
\end{equation}
Thus $B_\mu>0$ indicates that disturbance-first control recruits greater adaptive regulation across the ramp, while $B_{\Delta C}$ measures the corresponding ordering-dependent displacement in the state observable. The robustness plots report both the ensemble mean of these quantities and the fraction of matched pairs for which each $B_y$ is positive.

\subsection{Interpretive criteria}

Hysteresis is evaluated by separation between increasing-target and decreasing-target branches at the same $S^\ast$ and by replicate-level signed loop areas. An ordering-dependent regulatory-burden effect is evaluated by RF/DF separation in $\mu$, positive paired differences $D_\mu^{(b)}(S^\ast)$ at the reference condition, and positive $B_\mu$ across the local parameter grid. An ordering-dependent state-observable effect would require systematic RF/DF separation in $\Delta C$, a corresponding paired difference $D_{\Delta C}^{(b)}(S^\ast)$, or robust nonzero $B_{\Delta C}$ across conditions. These criteria distinguish controller effort from regulated state behavior.

\section{Results}

The results are organized around four questions. First, does adaptive gain show hysteresis under a continuous target-entropy ramp? Second, does causal ordering change the regulatory burden required to traverse the ramp? Third, does the state-level coherence-gap observable show the same ordering dependence? Fourth, do the main effects persist across a local neighborhood of disturbance amplitudes and initial gains?

\subsection{Adaptive gain exhibits robust carryover under continuous target modulation}

\begin{figure}[H]
    \centering
    \includegraphics[width=0.99\linewidth]{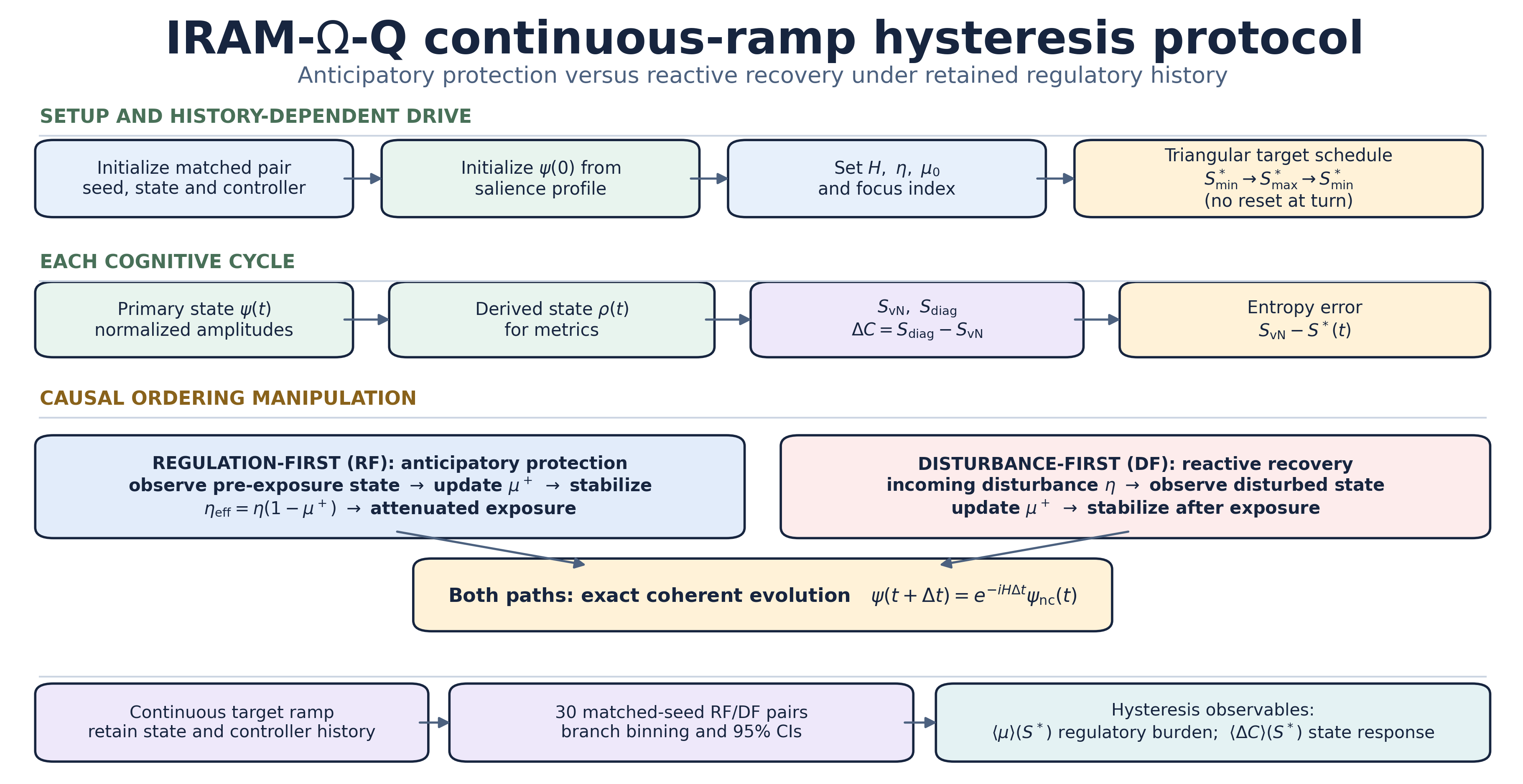}
    \caption{\textbf{Causal ordering and continuous-ramp analysis protocol.} Regulation-first (RF) applies anticipatory stabilization and attenuates incoming current-cycle disturbance before exposure, whereas disturbance-first (DF) permits incoming disturbance to act before reactive stabilization. Both paths subsequently undergo the same exact coherent evolution. The protocol applies a continuous triangular target-entropy ramp without resetting state or controller at the turning point, then compares replicate-averaged hysteresis in adaptive gain $\langle\mu\rangle(S^\ast)$ and coherence gap $\langle\Delta C\rangle(S^\ast)$ across 30 matched-seed RF/DF trajectory pairs.}
    \label{fig:causal_loop}
\end{figure}

Figure~\ref{fig:causal_loop} summarizes the causal ordering and continuous-ramp protocol. Regulation-first control applies stabilization before current-cycle disturbance exposure, while disturbance-first control permits incoming disturbance to act before reactive stabilization. Both orderings then undergo the same coherent evolution.

\begin{figure}[H]
    \centering
    \includegraphics[width=0.98\linewidth]{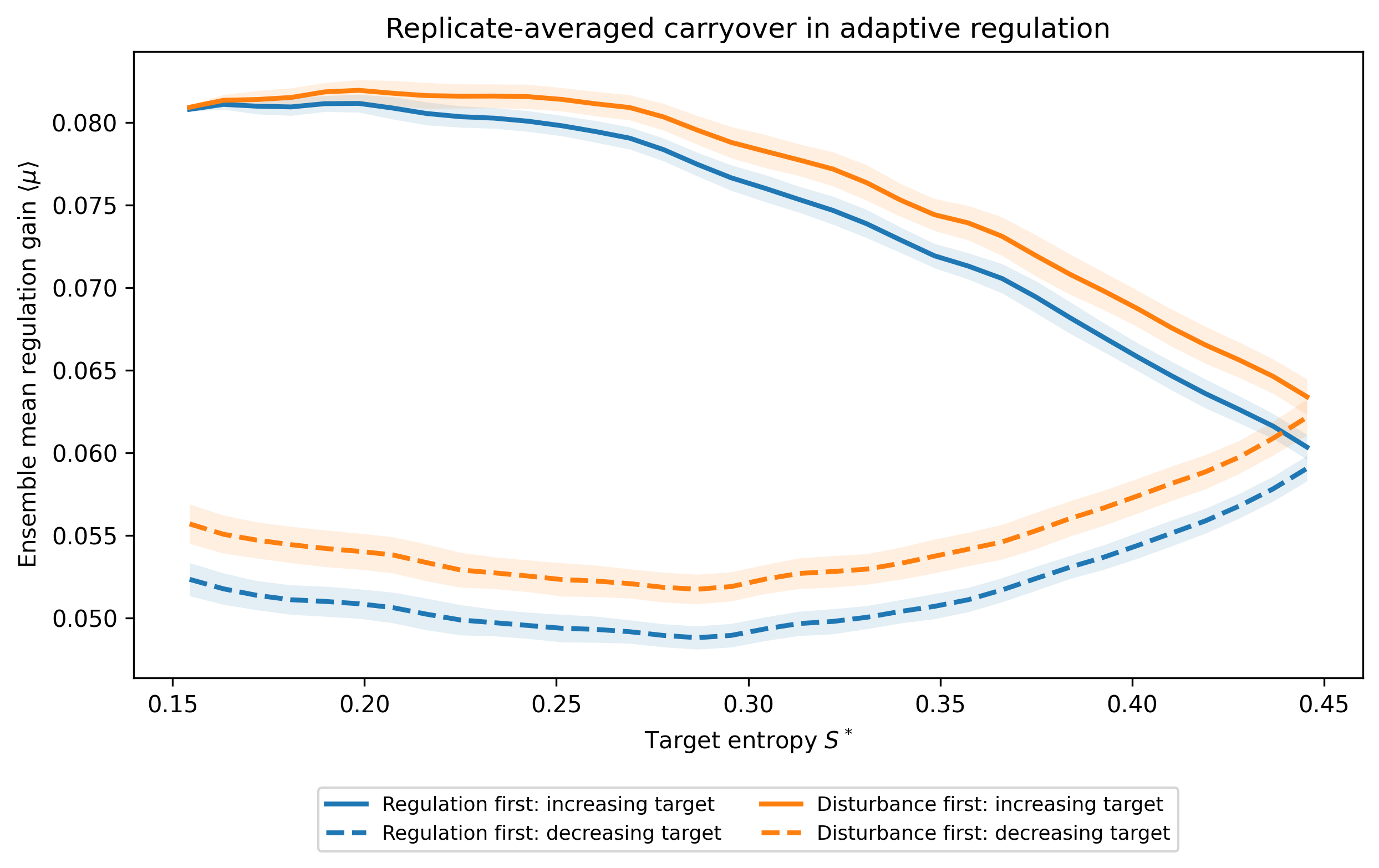}
    \caption{\textbf{Replicate-averaged carryover in adaptive regulation under a continuous target-entropy ramp.} Ensemble mean regulation gain $\langle\mu\rangle$ is plotted against target entropy $S^\ast$ for 30 matched-seed RF/DF trajectory pairs. Solid curves indicate the increasing-target branch and dashed curves indicate the decreasing-target branch; shaded regions show approximate 95\% confidence intervals across replicate-level binned trajectories. The strong separation between increasing and decreasing branches demonstrates hysteresis, while the systematically elevated DF curves indicate greater regulatory burden under disturbance-first ordering.}
    \label{fig:mu_hysteresis}
\end{figure}

Figure~\ref{fig:mu_hysteresis} shows replicate-averaged adaptive gain as a function of target entropy. The increasing-target and decreasing-target branches remain strongly separated across most of the target range. Thus, the gain recruited at a given value of $S^\ast$ cannot be predicted from the instantaneous target alone. It depends on whether the simulated agent is being driven toward higher or lower target entropy. This persistent branch separation after averaging 30 trajectories demonstrates robust hysteresis in adaptive regulation.

The loop has a clear dynamical interpretation. During the increasing-target branch, gain remains relatively high before falling as the schedule approaches its turning point. During the decreasing-target branch, gain follows a distinct lower return path rather than retracing the outward trajectory. Since no reset occurs at the turning point, the return branch reflects carryover from the state and controller history accumulated during ascent.

\subsection{Disturbance-first ordering carries greater adaptive control demand}

The same figure reveals a consistent ordering effect in regulatory demand. Across most of both the increasing-target and decreasing-target branches, the disturbance-first mean-gain curve lies above the corresponding regulation-first curve. Because paired trajectories share stochastic seeds and numerical settings, this separation indicates that reactive regulation recruits greater adaptive gain than anticipatory regulation under the continuous-ramp protocol.

This claim concerns regulatory demand rather than a direct judgment of state quality. Disturbance-first control is not labeled inferior simply because it has a different state trajectory. Rather, it generally requires a larger controller gain to operate through the same history-dependent target schedule. The result extends the causal-ordering analysis from fixed operating conditions to a dynamic carryover protocol: the advantage of earlier stabilization remains visible when the simulated agent is driven through a retained-history trajectory.

\subsection{Coherence gap exhibits history dependence with limited ordering separation}

\begin{figure}[H]
    \centering
    \includegraphics[width=0.98\linewidth]{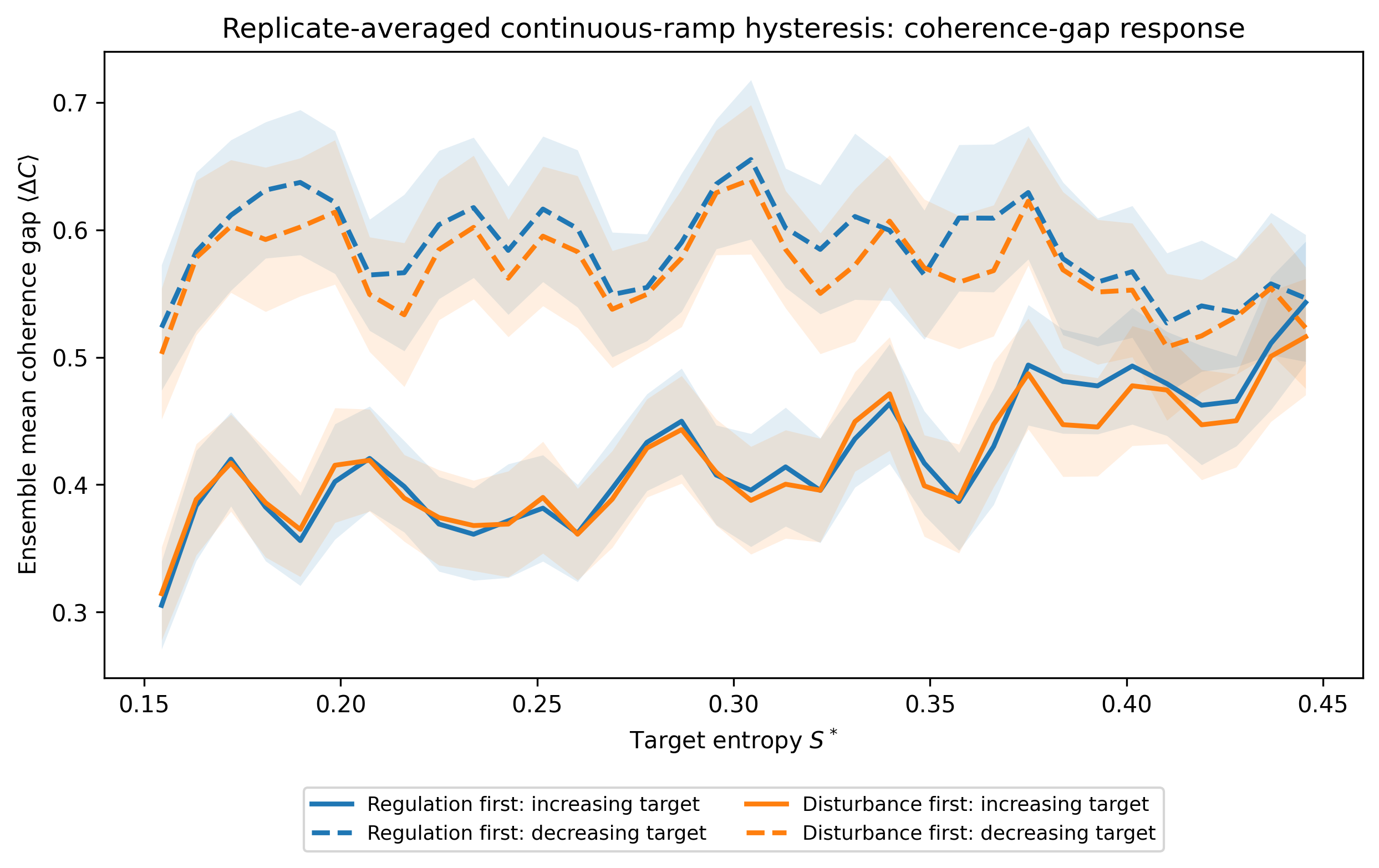}
    \caption{\textbf{Replicate-averaged coherence-gap response under a continuous target-entropy ramp.} Ensemble mean coherence gap $\langle\Delta C\rangle$ is plotted against target entropy $S^\ast$ for 30 matched-seed RF/DF trajectory pairs. Increasing-target and decreasing-target branches remain separated over much of the ramp, demonstrating path dependence in the state-level observable. RF and DF trajectories largely overlap within each branch, indicating that the principal ordering-dependent difference appears in adaptive gain rather than in the achieved coherence-gap response.}
    \label{fig:dc_hysteresis}
\end{figure}

Figure~\ref{fig:dc_hysteresis} shows the replicate-averaged coherence-gap response. Increasing-target and decreasing-target branches occupy separated response ranges over much of the ramp, with the decreasing branch systematically elevated relative to the increasing branch. Thus, the state-level observable also retains history: the same instantaneous target entropy is associated with a different mean coherence gap depending on the direction from which it is approached.

In contrast with the adaptive-gain figure, the ordering-specific coherence-gap curves within a given branch remain close, and their uncertainty bands largely overlap. Therefore, the coherence-gap result supports a claim of branch-dependent state history, but it does not support a strong ordering-dependent shift in the achieved state response under this protocol. The principal ordering effect is expressed in control effort: disturbance-first control maintains higher adaptive gain while producing broadly comparable branch-dependent coherence-gap behavior.

\subsection{Loop-area summaries quantify hysteresis across replicates}

\begin{figure}[H]
    \centering
    \includegraphics[width=0.98\linewidth]{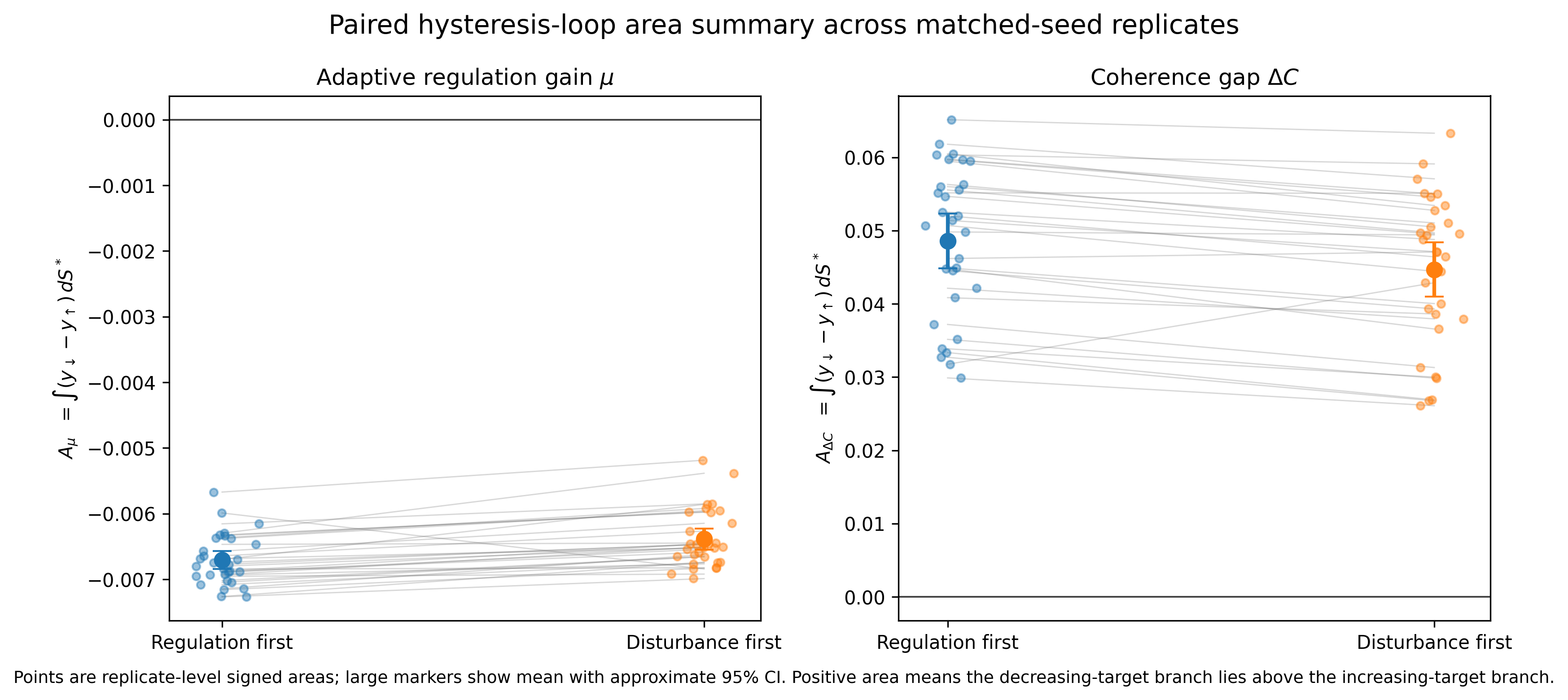}
    \caption{\textbf{Paired replicate-level summary of hysteresis-loop area.}
Signed loop areas are computed separately for adaptive gain and coherence gap. Small points joined by lines denote matched-seed RF/DF replicate pairs; large markers with error bars denote ensemble means with approximate 95\% confidence intervals. This plot quantifies the persistence and ordering dependence of carryover without relying only on visual separation of branch curves.}
    \label{fig:area_summary}
\end{figure}

Figure~\ref{fig:area_summary} converts branch separation into a replicate-level signed-area summary. Each paired trajectory contributes an area estimate for adaptive gain and for coherence gap. The displayed means and confidence intervals quantify whether loop orientation and magnitude persist across stochastic histories. The adaptive-gain area summarizes controller carryover, while the coherence-gap area summarizes state-observable carryover. This representation prevents the hysteresis conclusion from resting only on visual inspection of ensemble curves.

\subsection{Paired difference curves isolate ordering-dependent regulatory burden}

\begin{figure}[H]
    \centering
    \includegraphics[width=0.98\linewidth]{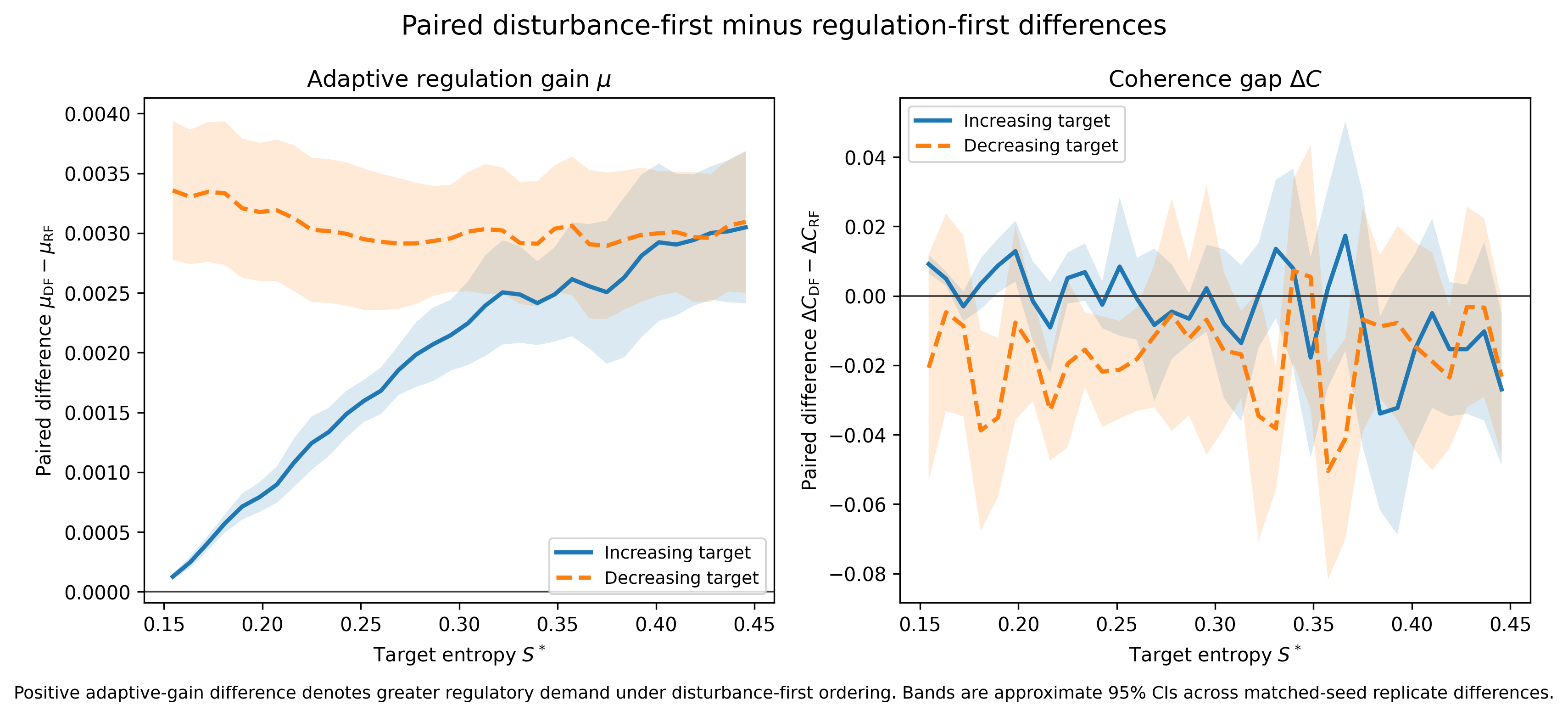}
    \caption{\textbf{Direct paired RF/DF difference across the continuous target ramp.} Curves show the matched-seed ensemble mean of DF minus RF separately on increasing-target and decreasing-target branches, with approximate 95\% confidence intervals across replicate-level paired differences. Positive $D_\mu$ denotes additional adaptive regulatory demand under disturbance-first ordering. The corresponding $D_{\Delta C}$ panel tests whether the controller-level ordering difference is accompanied by a comparable shift in the state observable.}
    \label{fig:paired_difference}
\end{figure}

Figure~\ref{fig:paired_difference} expresses the ordering comparison within matched pairs as disturbance-first minus regulation-first. For adaptive gain, the paired difference is positive over most of the continuous ramp on both branches. This supports the conclusion that disturbance-first control recruits greater adaptive gain than regulation-first control under otherwise matched histories.

The corresponding coherence-gap paired difference is comparatively small, consistent with the overlap of the ordering-specific state-observable trajectories in Figure~\ref{fig:dc_hysteresis}. Thus, the ordering effect is clearest as a difference in regulatory demand, rather than as a large difference in the attained coherence-gap response.

\subsection{Joint interpretation: comparable state behavior at different regulatory cost}

Together, the branch curves and paired summaries distinguish path dependence from control efficiency. Both $\mu$ and $\Delta C$ show separation between outward and return branches, demonstrating carryover under continuous target modulation. The paired ordering analysis makes the controller-level effect explicit: disturbance-first control requires higher adaptive gain than regulation-first control across most of the schedule, whereas the state-level ordering difference remains modest.

This pattern supports the interpretation that the two orderings navigate similar state-level history dependence, but disturbance-first control does so with a larger adaptive-regulation burden. When stabilization is applied before the disturbance step, current-cycle exposure is reduced before the state update is completed. When the full disturbance has already acted before recovery begins, the controller must respond after perturbation has entered the state. Under a slowly varying target schedule with retained history, that timing difference appears as persistent additional control demand in the reactive ordering rather than as a large displacement of the regulated coherence-gap trajectory.

\subsection{Parameter-neighborhood robustness of hysteresis and ordering effects}

\begin{figure}[H]
    \centering
    \includegraphics[width=0.99\linewidth]{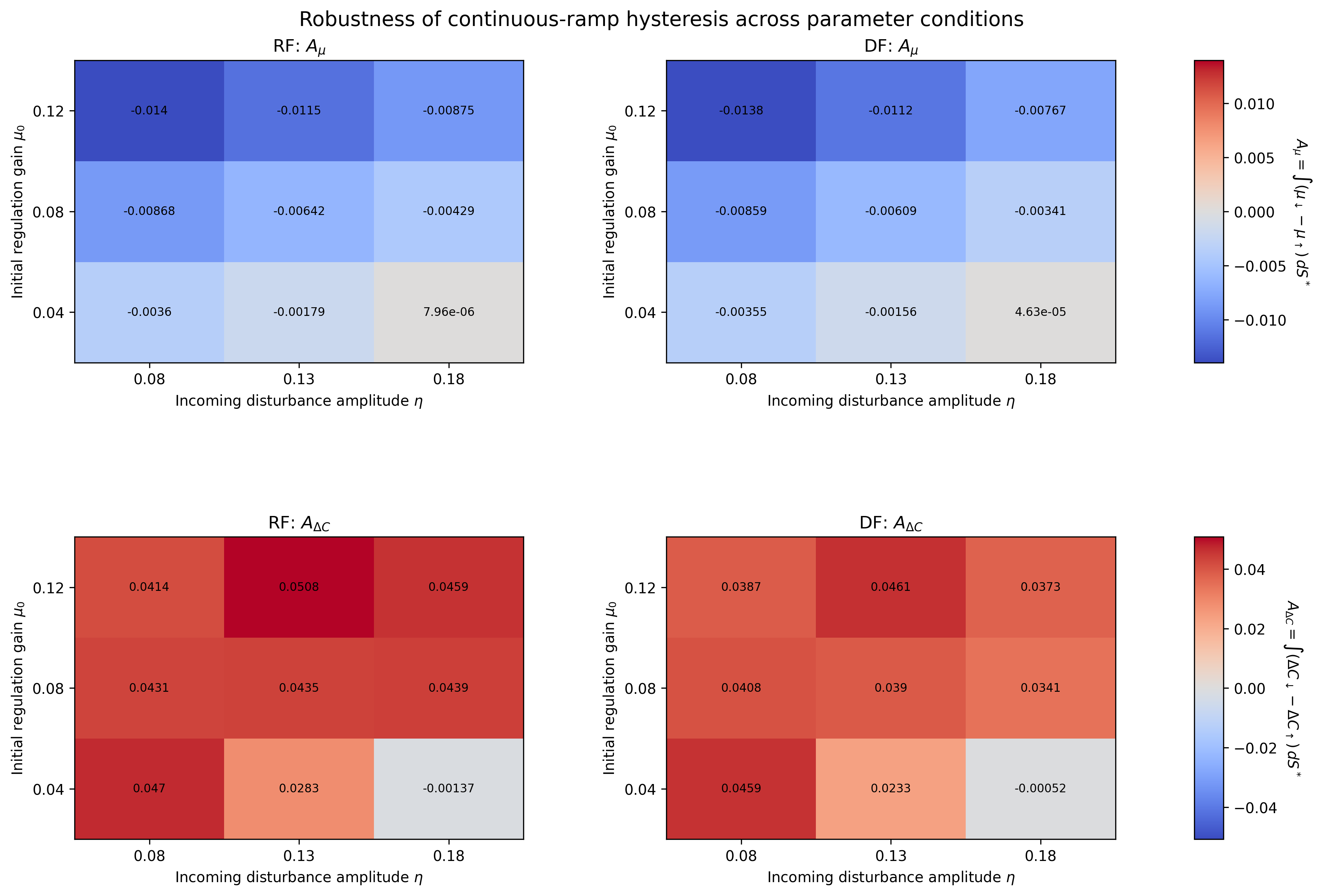}
    \caption{\textbf{Parameter-neighborhood robustness of continuous-ramp hysteresis.} Signed loop areas $A_\mu$ and $A_{\Delta C}$ are evaluated across incoming disturbance amplitudes $\eta\in\{0.08,0.13,0.18\}$ and initial regulation gains $\mu_0\in\{0.04,0.08,0.12\}$. Each cell summarizes 30 matched-seed RF/DF replicate pairs. RF and DF panels for a given observable share the same color scale; positive signed area indicates that the decreasing-target branch lies above the increasing-target branch.}
    \label{fig:robustness_areas}
\end{figure}

\begin{figure}[H]
    \centering
    \includegraphics[width=0.98\linewidth]{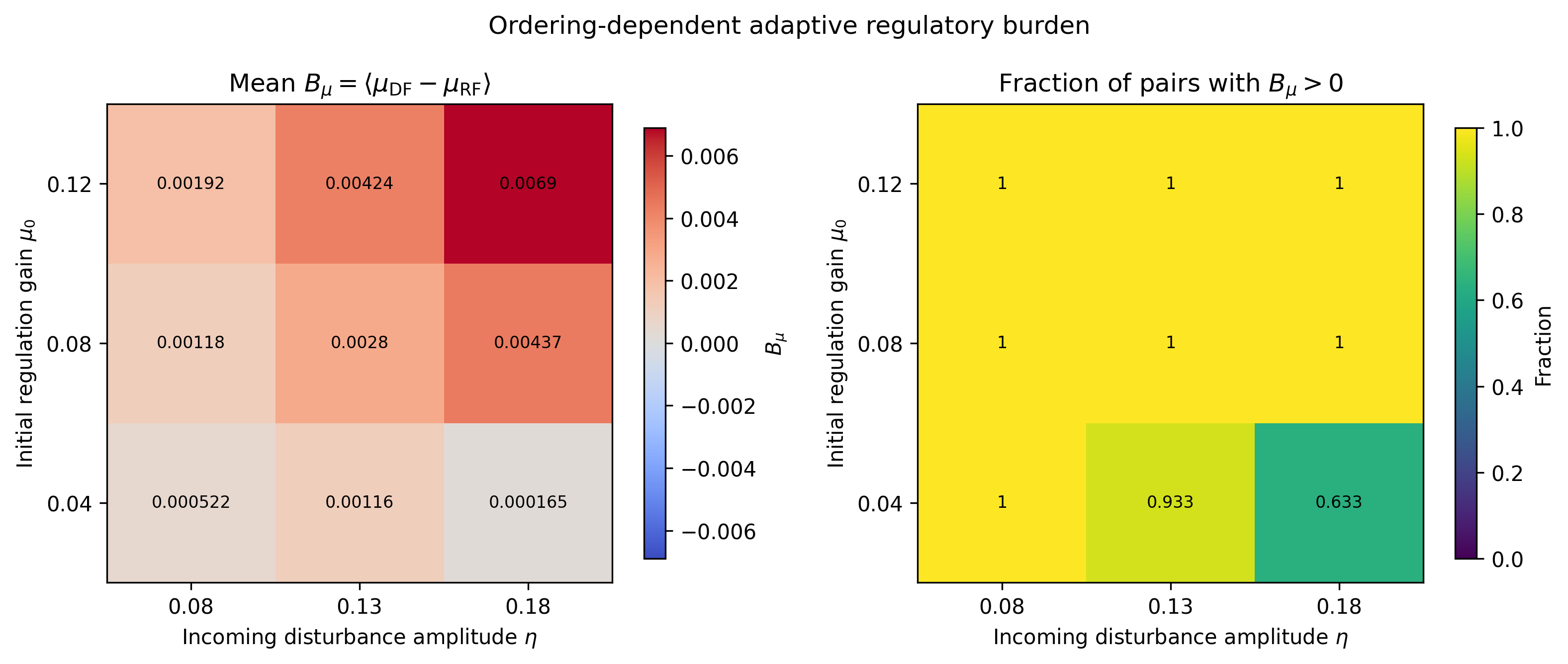}
    \caption{\textbf{Robustness of ordering-dependent adaptive regulatory burden.} The left panel reports the cellwise ensemble mean of $B_\mu=\langle\mu_{\mathrm{DF}}-\mu_{\mathrm{RF}}\rangle$ across both ramp branches and target-entropy bins. The right panel reports the fraction of matched RF/DF replicate pairs with $B_\mu>0$. Positive $B_\mu$ indicates greater adaptive regulation recruited under disturbance-first ordering.}
    \label{fig:robustness_mu}
\end{figure}

\begin{figure}[H]
    \centering
    \includegraphics[width=0.98\linewidth]{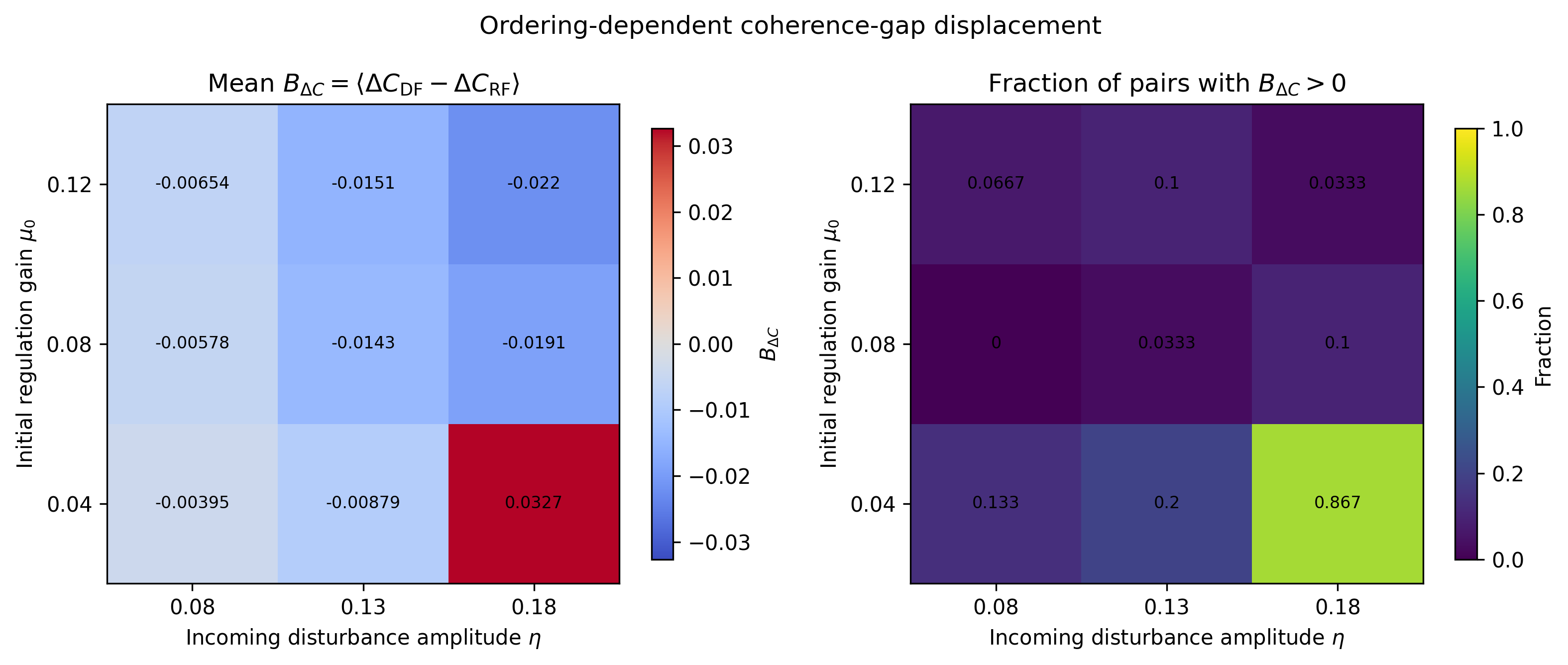}
    \caption{\textbf{Robustness of ordering-dependent coherence-gap displacement.} The left panel reports the cellwise ensemble mean of $B_{\Delta C}=\langle\Delta C_{\mathrm{DF}}-\Delta C_{\mathrm{RF}}\rangle$ across both ramp branches and target-entropy bins. The right panel reports the fraction of matched RF/DF replicate pairs with $B_{\Delta C}>0$. This figure tests whether an ordering-dependent regulatory burden is accompanied by a systematic change in the state-level coherence-gap observable.}
    \label{fig:robustness_dc}
\end{figure}

Figures~\ref{fig:robustness_areas}--\ref{fig:robustness_dc} extend the reference-condition analysis across nearby disturbance amplitudes and initial regulation gains. Figure~\ref{fig:robustness_areas} shows that the adaptive-gain loop area $A_\mu$ is similar under regulation-first and disturbance-first timing throughout the tested grid. The loop is negative in eight of nine conditions for both orderings, reflecting a lower-gain decreasing-target branch under the present sign convention, and approaches zero only in the lowest-gain, highest-disturbance condition $(\mu_0,\eta)=(0.04,0.18)$. Thus, the adaptive-gain hysteresis observed at the reference condition is not an isolated feature of a single parameter choice.

The stronger ordering result appears in the paired burden statistic shown in Figure~\ref{fig:robustness_mu}. The cellwise mean
\[
B_\mu=\langle\mu_{\mathrm{DF}}-\mu_{\mathrm{RF}}\rangle
\]
is positive in all nine parameter conditions, ranging from approximately $1.65\times10^{-4}$ to $6.90\times10^{-3}$. In eight of nine cells, at least $93.3\%$ of matched pairs have $B_\mu>0$, and in six cells all 30 paired replicates are positive.

The weakest consistency occurs at the low-initial-gain/high-disturbance boundary condition $(\mu_0,\eta)=(0.04,0.18)$. This cell combines the smallest tested initial regulatory authority with the strongest tested disturbance, so the controller begins the ramp with limited capacity to attenuate incoming perturbation. In this regime, the mean regulatory-burden displacement $B_\mu$ remains positive, but the fraction of matched pairs with $B_\mu>0$ falls to $63.3\%$, indicating that the regulation-first advantage becomes less consistent across stochastic histories.

The coherence-gap analysis reveals a more parameter-dependent pattern. Figure~\ref{fig:robustness_areas} shows positive $A_{\Delta C}$ in eight of nine cells under each ordering, demonstrating persistent state-level branch separation across most of the parameter neighborhood. However, Figure~\ref{fig:robustness_dc} does not support a universal ordering effect at the state-observable level. The paired displacement
\[
B_{\Delta C}=\langle\Delta C_{\mathrm{DF}}-\Delta C_{\mathrm{RF}}\rangle
\]
is negative in eight cells, ranging from approximately $-3.95\times10^{-3}$ to $-2.20\times10^{-2}$, but changes sign at $(\mu_0,\eta)=(0.04,0.18)$, where it is positive and larger in magnitude, $3.27\times10^{-2}$. Accordingly, $\Delta C$ supports robust hysteresis and local ordering sensitivity, but not a simple parameter-independent claim that one ordering uniformly lowers the coherence-gap response.

Table~\ref{tab:robustness_summary} collects the principal quantitative robustness outcomes. The uniform sign of $B_\mu$ across the tested grid contrasts with the parameter-dependent sign of $B_{\Delta C}$, reinforcing the conclusion that the most stable ordering effect is a reduction in adaptive regulatory burden under regulation-first timing rather than a universal displacement of the coherence-gap observable.

\begin{table}[H]
\centering
\caption{\textbf{Compact numerical summary of the continuous-ramp robustness analysis.}
The paired burden statistic $B_y$ is the mean DF-minus-RF difference across ramp branches and target-entropy bins. Positive $B_\mu$ indicates greater adaptive regulation recruited under disturbance-first ordering.}
\label{tab:robustness_summary}
\begin{tabular}{p{0.50\linewidth}p{0.38\linewidth}}
\hline
Quantity & Result \\
\hline
Parameter neighborhood examined
    & $\eta\in\{0.08,0.13,0.18\}$,
      $\mu_0\in\{0.04,0.08,0.12\}$ \\[2pt]

Matched-seed replicate pairs per parameter cell
    & $30$ \\[2pt]

Total matched RF/DF pairs in robustness grid
    & $270$ pairs ($540$ trajectories) \\[2pt]

Cells with positive mean regulatory-burden displacement,
$B_\mu>0$
    & $9/9$ cells \\[2pt]

Range of mean $B_\mu$
    & $1.65\times10^{-4}$ to
      $6.90\times10^{-3}$ \\[2pt]

Cells with at least $93.3\%$ of paired replicates satisfying
$B_\mu>0$
    & $8/9$ cells \\[2pt]

Weakest regulatory-burden consistency condition
    & $(\mu_0,\eta)=(0.04,0.18)$;
      positive mean $B_\mu$, $63.3\%$ positive pairs \\[2pt]

Cells with positive coherence-gap hysteresis area
$A_{\Delta C}$ under each ordering
    & $8/9$ cells for RF and $8/9$ cells for DF \\[2pt]

Cells with negative mean coherence-gap displacement,
$B_{\Delta C}<0$
    & $8/9$ cells \\[2pt]

Sign-reversal condition for $B_{\Delta C}$
    & $(\mu_0,\eta)=(0.04,0.18)$;
      $B_{\Delta C}=3.27\times10^{-2}$ \\[2pt]
\hline
\end{tabular}
\end{table}

\section{Discussion}

The present analysis tests whether ordering effects observed under fixed conditions persist when the simulated agent retains its state and controller history across a continuous target-ramp protocol. 
The adaptive-gain loop is the principal result: hysteresis is stable after replicate averaging, and the ordering effect survives across matched stochastic histories. The evidence therefore supports a control-efficiency interpretation of regulation-first timing. Anticipatory placement of regulation allows the agent to traverse the same continuous target schedule while generally recruiting less adaptive gain than a reactive disturbance-first sequence.

The coherence-gap result adds an important qualification. State-level history dependence is evident at the reference condition and remains present in most nearby parameter conditions, but its RF/DF ordering effect is not uniform. This prevents an overstatement that regulation-first timing broadly lowers all dynamical measures. Instead, the model distinguishes two questions: whether the internal state carries history and whether one causal ordering sustains that behavior at lower regulatory effort. The first is supported by branch separation in $\Delta C$; the second is supported much more uniformly by the positive $B_\mu$ result.

This distinction matters for adaptive-agent design. Two agents may look similar in state or behavior while differing in the amount of regulation needed to remain stable. A controller that waits until after disturbance exposure may still keep the system organized, but at greater adaptive cost. In a physical system, that cost might appear as stronger corrections, increased actuator activity, or tighter filtering. In a computational system, it might appear as additional sampling, more optimization steps, higher latency, or greater internal compute. The present model abstracts over these implementations and studies the common control-demand structure.

The results are also relevant to cognitively inspired interpretations of regulation. Anticipatory stabilization and reactive recovery need not differ primarily in their final state. They may differ in effort, timing, and accumulated burden. In that sense, the model provides a computational way to separate state quality from regulatory demand. Efficient regulation is not simply a matter of reaching a target; it is also a matter of how much corrective effort is required to remain near it under disturbance and retained history.

\section{Limitations}

The present simulations isolate internal regulatory dynamics rather than embedding the agent in an external task benchmark. Adaptive gain is therefore interpreted as modeled control demand, not as a task-performance score.

The robustness study tests a local neighborhood around the reference condition. Its purpose is to check that the ordering effect is not an artifact of a single parameter choice.

\section{Conclusion}

When the simulated agent is driven through a continuous target-entropy ramp, its regulatory dynamics show robust history dependence. Adaptive gain forms a reproducible hysteresis loop across 30 matched-seed regulation-first/disturbance-first trajectory pairs, showing that the regulation required at a given target depends on the trajectory by which the agent arrived there.

The central ordering result is focused and consistent: disturbance-first control recruits greater adaptive gain than regulation-first control across the tested conditions. Across all nine tested combinations of disturbance amplitude and initial regulation gain, the mean paired regulatory-burden statistic satisfies $B_\mu>0$. The coherence-gap observable also displays branch-dependent history across most of the tested neighborhood, but its ordering displacement is not uniform and reverses sign in the low-initial-gain/high-disturbance corner.

Anticipatory regulation does not uniformly shift every state-level measure. Its clearest effect is on regulatory cost: the simulated agent traverses the same path-dependent protocol with lower adaptive gain while maintaining broadly comparable coherence-gap behavior. In this sense, causal timing changes the burden of regulation. Stabilizing before disturbance exposure allows the agent to remain organized with less adaptive effort than recovering after the disturbance has already acted.

\section*{Data and code availability}

The simulations were performed with the author's IRAM-$\Omega$-Q codebase. Numerical summaries supporting the reported figures are available from the author upon reasonable request. The simulation codebase is maintained separately from the manuscript.

\end{document}